\documentclass[twocol]{ametsocV6.1}

\title{Pixels and Predictions: Potential of GPT-4V in Meteorological Imagery Analysis and Forecast Communication}

\authors{
John R. Lawson,\aff{a,b}\correspondingauthor{John R. Lawson, john.lawson@usu.edu}
Joseph E. Trujillo-Falc\'on\thanks{Current affiliation of Trujillo-Falc\'on: Dept. of Climate, Meteorology \& Atmospheric Sciences; and Dept. of Communication; University of Illinois Urbana-Champaign, Urbana, Illinois},\aff{c,d,e} 
David M. Schultz,\aff{f,g} 
Montgomery L. Flora,\aff{c,d,h}
Kevin H. Goebbert,\aff{i}
Seth N. Lyman,\aff{a,j}
Corey K. Potvin,\aff{c,h,k}
Adam J. Stepanek\aff{i}
}

\affiliation{
\aff{a}{Bingham Research Center, Utah State University, Vernal, Utah}\\
\aff{b}{Dept. of Mathematics and Statistics, Utah State University, Logan, Utah}\\
\aff{c}{NOAA/OAR/National Severe Storms Laboratory, Norman, Oklahoma}\\
\aff{d}{Cooperative Institute for Severe and High-Impact Research and Operations, The University of Oklahoma, Norman, Oklahoma}\\
\aff{e}{Department of Communication, The University of Oklahoma, Norman, Oklahoma}\\
\aff{f}{Centre for Crisis Studies and Mitigation, University of Manchester, Manchester, United Kingdom}\\
\aff{g}{Department of Earth and Environmental Sciences, University of Manchester, Manchester, United Kingdom}\\
\aff{h}{NSF AI Institute for Research on Trustworthy AI in Weather, Climate, and Coastal Oceanography, Norman, Oklahoma}\\
\aff{i}{Department of Geography and Meteorology, Valparaiso University, Valparaiso, Indiana}\\
\aff{j}{Department of Biochemistry and Chemistry, Utah State University, Logan, Utah}\\
\aff{k}{School of Meteorology, The University of Oklahoma, Norman, Oklahoma}
}
\thanks{This manuscript has been submitted to \textit{Artificial Intelligence for the Earth Systems}. Copyright in this manuscript may be transferred without further notice. Supplementary material (Fig. S1) is found at the end of this manuscript.}

\usepackage[utf8]{inputenc}
\usepackage[spanish,english]{babel}

\usepackage{pdfpages}

\abstract{Generative AI, such as OpenAI's GPT-4V large-language model, has rapidly entered mainstream discourse. Novel capabilities in image processing and natural-language communication may augment existing forecasting methods. Large language models further display potential to better communicate weather hazards in a style honed for diverse communities and different languages. This study evaluates GPT-4V's ability to interpret meteorological charts and communicate weather hazards appropriately to the user, despite challenges of \textit{hallucinations}, where generative AI delivers coherent, confident, but incorrect responses. We assess GPT-4V's competence in two tasks: (1) generating a severe-weather outlook from weather-chart analysis and conducting self-evaluation, revealing an outlook that corresponds well with a Storm Prediction Center human-issued forecast; and (2) producing hazard summaries in Spanish and English from weather charts. Responses in Spanish, however, resemble direct (not idiomatic) translations from English to Spanish, yielding poorly translated summaries that lose critical idiomatic precision required for optimal communication. Our findings advocate for cautious integration of tools like GPT-4V in meteorology, underscoring the necessity of human oversight and development of trustworthy, explainable AI.}

\begin{document}

\maketitle

%
%
%
\statement The integration of generative AI such as GPT-4V into meteorology brings opportunity for improving forecast communication of weather hazards across languages and communities. This study evaluates GPT-4V's capacity to create plausible severe-weather outlooks, and communicate hazards in both Spanish and English, from inputs of weather charts and text. The weather outlook generally aligns with a human-generated forecast product but displays vagueness and incorrect reasoning. Further, translations lack idiomatic precision and display poor grasp of cultural nuance. Despite this, GPT-4V shows potential for advancement in meteorological application. We advocate for cautious AI integration, emphasizing need for human oversight and reliable, trustworthy output. 

%
%
%


\section{Introduction}
Throughout the information age, scientists have embraced computers as allies---from automation of simple tasks to post-processing big datasets. Human weather forecasters provide added value over raw numerical weather prediction (NWP) guidance, both quantitatively \citep{Novak2011-eq} and in public communication \citep{Stuart2006-st}. Today, generative AI is rapidly entering the mainstream in the wake of intuitive web-based ``chatbot" interface, such as OpenAI's ChatGPT large-language model (\url{chat.openai.com}, accessed 1 July 2024). Within this flurry of AI-model development, meteorologists now hold unprecedented but little-explored tools for improving the weather-forecasting enterprise. The addition of ``sight" to language models---so-called multi-modal models that can ingest more than text---enables weather images and charts to be interpreted not just by the human eye but by machine intelligence. Given this combination of natural-language processing, intelligence, and ingestion of imagery, we ask: \textit{Can this rapid technological advance serve meteorologists in forecasting, research, and public communication?} We do not seek to replace the human forecaster; rather, we evaluate competence of the first public release of GPT-4V to augment the meteorologist's toolbox. 

In the era of big data, scientists are increasingly turning to multi-modal AI models like GPT-4V to increase efficiency and achieve more in less time. Recent AI models can perform specific tasks as well as humans \citep{Bubeck2023-ls,Yang2023-oy} and may disrupt society more than the last generation of LLMs such as GPT-3 \citep{Floridi2020-hd,Tamkin2021-hg,Bender2021-rg}. Yet, despite typically coherent and confident responses to prompts, there is the risk of \textit{hallucinations}: instances where output confidently and convincingly contains erroneous information. \citet{Frankfurt2005-jv} and \citet{Hicks2024-bf} eccentrically detail the distinction between hallucinations and indifference of some LLMs to truthfulness of their output. The language model generates plausible output based on patterns in the training corpus, but does not fact-check its inferences out-of-the-box. As the human developer or user, it is difficult to diagnose sources of error in proprietary LLMs when little information about training data is given. Even with the corpus dataset in hand, we estimate from \citet{Brown2020-nr} and \citet{Kaplan2020-js} that GPT-4V learned from 1--2\,TB of filtered text and an order of magnitude more in image processing. The forecast of severe weather is a risk-averse endeavor; small errors may have disproportionally negative consequences. Given this risk sensitivity and the importance of specific wording when communicating weather hazards to the public \citep{Rothfusz2018-yk,Trujillo-Falcon2021-np}, AI language models present potential to improve public communication of hazard risks \citep{Olteanu2014-nz} tailored to the audience.

These LLMs are trained on massive datasets of text and code, enabling them to perform a variety of tasks, including translation, summary of large chunks of text, and knowledge of a wide range of topics allowing tailored answering of questions (prompts). During the writing of the present manuscript, OpenAI have demonstrated further abilities, such as more natural interaction via speech (\textit{GPT-4o}) and text-to-video (\textit{Sora}). The GPT-4V system card \citep{OpenAI2023-lx} details the testing and evaluation performed by OpenAI, but the lack of transparency makes it difficult to assess performance through a report card. The reader is encouraged to consult \url{openai.com/index/gpt-4v-system-card} (accessed 1 August 2024) for more thorough testing. Our goal is to discern whether this somewhat nebulous skill-set within GPT-4V can constructively contribute to meteorological applications; we probe its capabilities and limits in meteorological application through three questions:

\begin{enumerate}
    \item Can GPT-4V correctly interpret weather charts and imagery?
    \item Can GPT-4V effectively communicate weather hazards in language tailored for the audience?
    \item What constitutes a useful answer -- what are our expectations and are they reasonable?
\end{enumerate}

\section{Method}
We choose GPT-4 as our large-language model due to OpenAI's claimed GPT-4V performance metrics and release of the ``vision" ability enabling image inputs. Further, at the time of writing, keyword searches on Google for ChatGPT and OpenAI were an order of magnitude larger (\url{trends.google.com}, accessed 15 March 2024; not shown) than competitors and their AI models such as Anthropic (\textit{Claude}) and Google (\textit{Gemini} or \textit{Bard}). Herein, we used the ChatGPT web-portal: the online graphical front-end to OpenAI GPT models. 
 
During initial exploration, many types of challenges were provided to GPT-4V; here, we focus on two tasks that are most representative of GPT-4V's range and aptitude of abilities. First, we ask GPT-4V to interpret multiple weather charts and deduce the risk of meteorological hazards. Second, we give GPT-4V a synoptic-scale forecast chart marked with regions of general weather type such as thunderstorms and request plain-language summaries in both Spanish and English. The conversations were processed by the lead author within ChatGPT, performed within two weeks of 1 October 2023 with the same version released in stages to the public starting on September 25 (\url{https://help.openai.com/en/articles/6825453-chatgpt-release-notes}, accessed 1 November 2023). We used no custom instructions, which can be provided to personalize responses within ChatGPT. We include sections of our GPT-4V conversations in the manuscript, but full conversations (unabridged other than trimming of unimportant procedural text) are found in the Supplementary Material. Text quoted verbatim from GPT-4V conversation is italicized. Our experiments were conducted within a closed environment (i.e., ChatGPT did not have access to the internet). The training corpus did not include information beyond September 2021.

\section{GPT-4V and Meteorological Prediction}
\citet{Yang2023-oy} and \citet{Bubeck2023-ls} showed that GPT-4 is able to interpret and conceptualize data spatially, such as text-based navigation after a description of a location. Indeed, \citet{Xu2024-wr} found LLMs displayed ability to hold large batches of maps in memory to form basic spatial awareness, but this was tempered by low reliability of image identification and black-box behaviour that precluded reproducible evaluation analysis. GPT-4V displays nascent ability to anticipate, which is a key human characteristic \citep{Dennett2015-xv}. Combining both factors, can GPT-4V conceptualize the atmospheric state from a sequence of charts?

\subsection{Initial set of weather charts}
We want to determine if GPT-4V can grasp the three-dimensional atmospheric flow, and hence show GPT-4V multiple pairs of charts depicting North American Model (NAM) and Global Forecasting System (GFS) guidance data, visualized by Pivotal Weather (\url{https://home.pivotalweather.com/}, accessed 15 October 2023). We provide two models to assist GPT-4V capture basic characteristics of uncertainty. While this is smaller corpus of guidance that would be available to human forecasters, we give sufficient information to GPT-4V that a human would capture the general flow pattern from the same resources:

\begin{itemize}
  \item Geopotential height at 300 hPa and 500 hPa
  \item Dry-bulb temperature at 500 hPa, 850 hPa, and 2 m
  \item Wind at 300 hPa, 500 hPa, 850 hPa, and 10 m
  \item Mean sea-level pressure
  \item Equivalent potential temperature at 2 m
  \item Simulated composite reflectivity
\end{itemize}

\clearpage

\begin{figure}[t]
  \centering
  \noindent\includegraphics[width=0.9\textwidth]{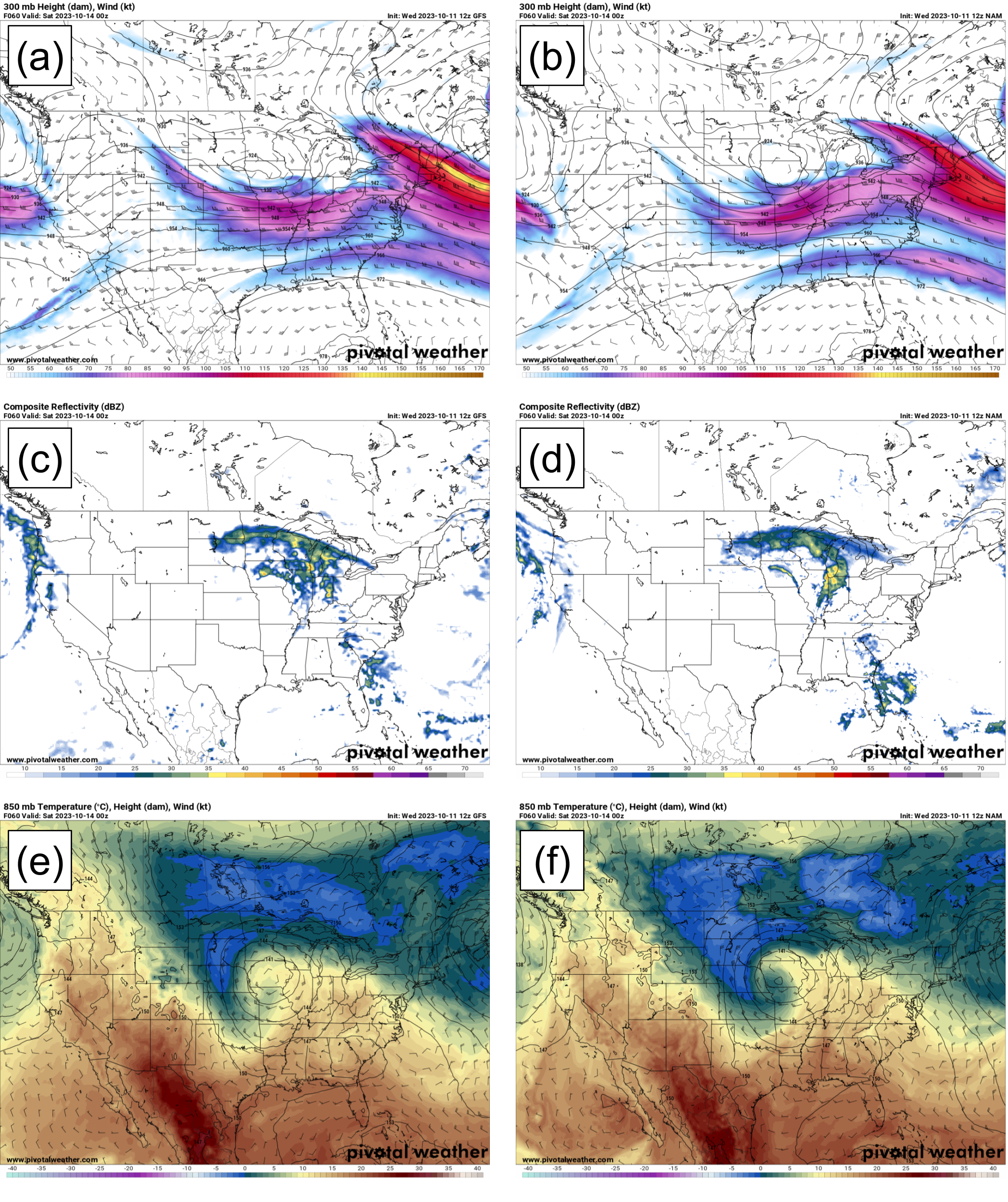}\\
  \caption{Three pairs of weather charts taken from a full set given to GPT-4V. Images reproduced with kind permission of Pivotal Weather, LLC. Geopotential height and wind speed at 300\,hPa (a,b); simulated composite reflectivity (c,d); dry-bulb temperature at 850\,hPa (e,f). Left column is GFS (a,c,e); right column is NAM (b,d,f) }\label{fig:spc_maps_1}
\end{figure}

\clearpage

\begin{figure}[t]
  \centering
  \noindent\includegraphics[width=0.9\textwidth]{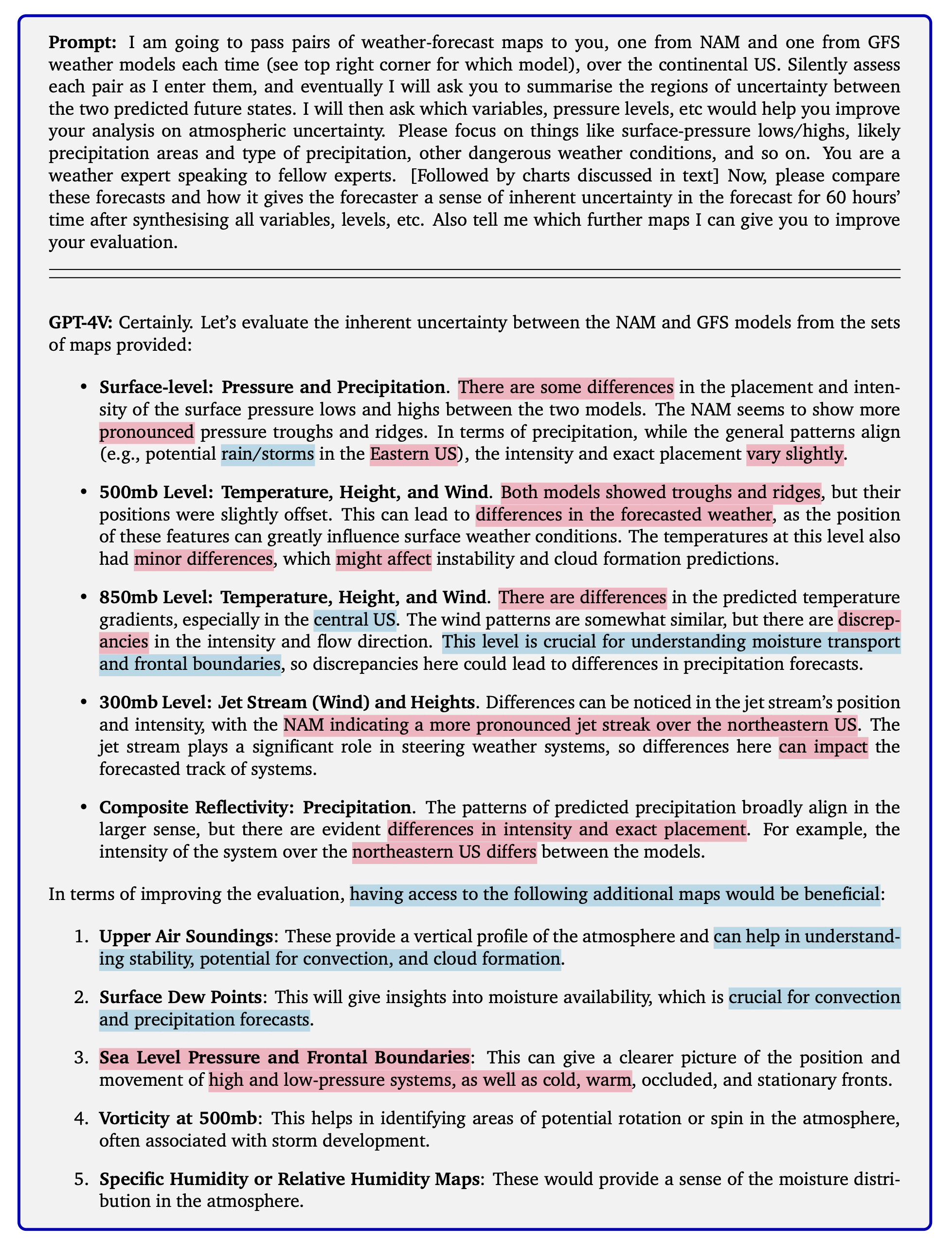}\\
  \caption{GPT-4V response to a collection of weather charts. We subjectively highlight vague/incorrect responses in red and useful/correct sections in blue.}\label{fig:spc_response_1}
\end{figure}

\clearpage

Figure~\ref{fig:spc_maps_1} shows a subset of these charts given to GPT-4V, including geopotential height and wind vectors on multiple levels critical for understanding the flow pattern. The forecast synoptic-scale flow pattern displays a vertically stacked extratropical cyclone over the U.S. Great Plains on the polar side of a jet stream (Fig.~\ref{fig:spc_maps_1}a,b) and evident in simulated composite reflectivity (Fig.~\ref{fig:spc_maps_1}c,d). The warm and cold sectors of the cyclone are seen in 850-hPa temperature (Fig.~\ref{fig:spc_maps_1}e,f). Our first instructional prompt after providing charts (Fig.~\ref{fig:spc_response_1}) asks GPT-4V to give a summary of uncertainty after the first batch of maps and how it might improve its conceptualization of the atmospheric state with further charts. Much of the response is not specific enough to provide utility to end users (Fig~\ref{fig:spc_response_1}), referring to \textit{some differences}, \textit{slight} variations, \textit{differences in forecasted weather} that might affect predictions. 

Responses display hallucinations falsely detected in the provided charts: Fig.~\ref{fig:spc_response_1} refers to precipitation in the Eastern US not shown in the charts (Fig.~\ref{fig:spc_maps_1}c,d). The choice of terminology is often non-standard: GPT-4V identifies a more \textit{pronounced} jet streak over the northeastern US. If we interpret ``\textit{pronounced}'' as meaning stronger in magnitude, this is evidently a mix-up between the two models, despite GPT-4V’s ability to read small text in images (not shown). It is unlikely a forecaster would find utility in GPT-4V's responses in Fig.~\ref{fig:spc_response_1}: vague replies require the human to continue their task by fetching further guidance on top of fact-checking the responses. However, a human forecaster would also have access to many more meteorological charts; accordingly, we now heed the request to provide addition chart(s).

\subsection{Second set of weather charts}

At the end of the response in Fig.~\ref{fig:spc_response_1}, GPT-4V requests five further pairs of charts to better grasp atmospheric uncertainty and structure:

\begin{itemize}
	\item Upper-air soundings, to assist understanding of stability and convective potential;
	\item Surface dew-point temperature (implying GPT-4V has not processed the 2-m equivalent-potential-temperature charts);
	\item Sea-level pressure and analyzed surface frontal boundaries (something GPT-4V can attempt to infer from the provided charts);
	\item Vorticity at 500 hPa (again, GPT-4V might infer this from existing upper-level charts);
	\item Humidity charts (at unspecified pressure levels).
\end{itemize}

The request for humidity data is sensible from GPT-4V's lack of access to upper-level humidity information; we give GFS and NAM charts of 700-hPa relative humidity and wind vectors, then request GPT-4V to infer frontal positions from these and previous charts. When asked about the equivalent-potential-temperature charts, given initially to improve understanding of near-surface cyclone structure (not shown), GPT-4V is not able to recall these. It is unclear whether this is due to these charts residing so deep in the conversation memory that this chart is deprioritized: GPT-4V will forget information if the model architecture deems it less relevant to imminent or recent prompts \citep{Vaswani2017-jw,OpenAI2023-lx}. Continuing its response, GPT-4V makes numerous mistakes, such as another conflation of NAM and GFS 300-hPa wind maxima and identifying discrepancies indiscernible to human eye in the Great Lakes region of the 700-hPa humidity charts (not shown; S1). GPT-4V's response also suggests a warm front north of the high-humidity region in the Great Lakes, but mistakenly locates a cold front in the northeastern US rather than stretching from Ohio to the Carolinas. 

The errors and lack of useful information in these responses early in the conversation may cause a chain-reaction of useless or harmful responses if not caught by a human quickly enough in responses. This failure to check self-consistency stems from lack of hybrid fast-and-slow thought process \citep{Kahneman2011-dn} to oversee fidelity: a known limitation of AI in emulating human thought and decision-making \citep{Booch2021-qp,Weston2023-zp}. \citet{Bubeck2023-ls} found when deriving or processing mathematical logic, where a statement is often either correct or incorrect, GPT-4V can reach a correct answer despite generating contradictory rationale. Similarly, as we progress through the conversation, previous logical and factual errors do not preclude a useful final answer---self-inconsistency does not preclude a course-correction. As such, we turn to GPT-4V's ability to recognize meteorological hazards in the charts provided, and find GPT-4V indeed has a reasonable grasp of the atmospheric state despite prior error. This further accents the need for humans to persist with their GPT-4V conversation rather than halting after one-shot tasks, partly due to low interpretability of response variation from a black-box model (e.g., \citealt{Flora2024-de}).

\clearpage

\begin{figure}[t]
  \centering
  \noindent\includegraphics[width=0.88\textwidth]{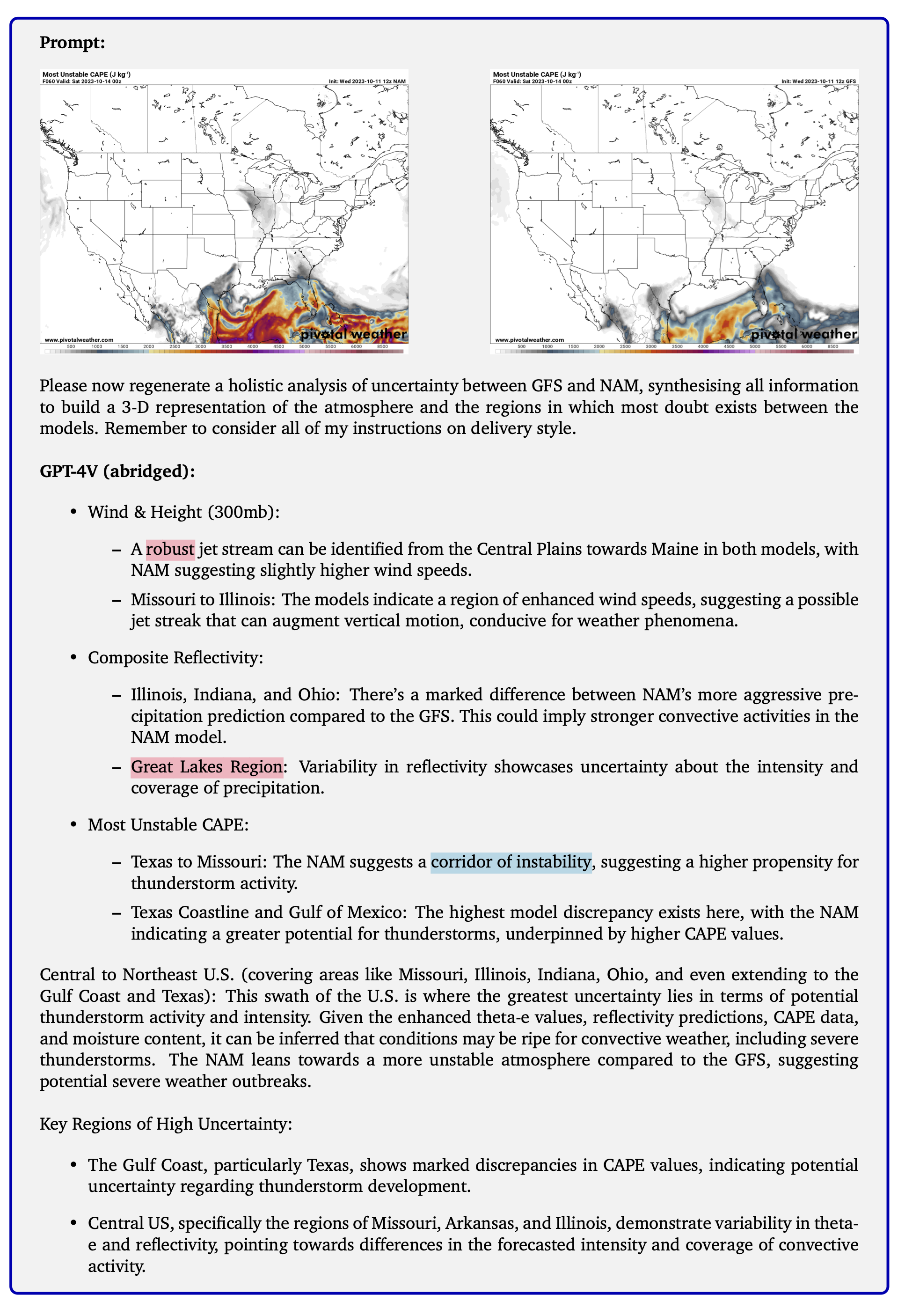}\\
  \caption{Conversation snippet with (a) MUCAPE maps and request, (b) response. We have removed discussion of maps not shown in Fig.~\ref{fig:spc_maps_1}.}\label{fig:spc_both}
\end{figure}

\clearpage

\begin{figure}[t]
  \centering
  \noindent\includegraphics[width=0.95\textwidth]{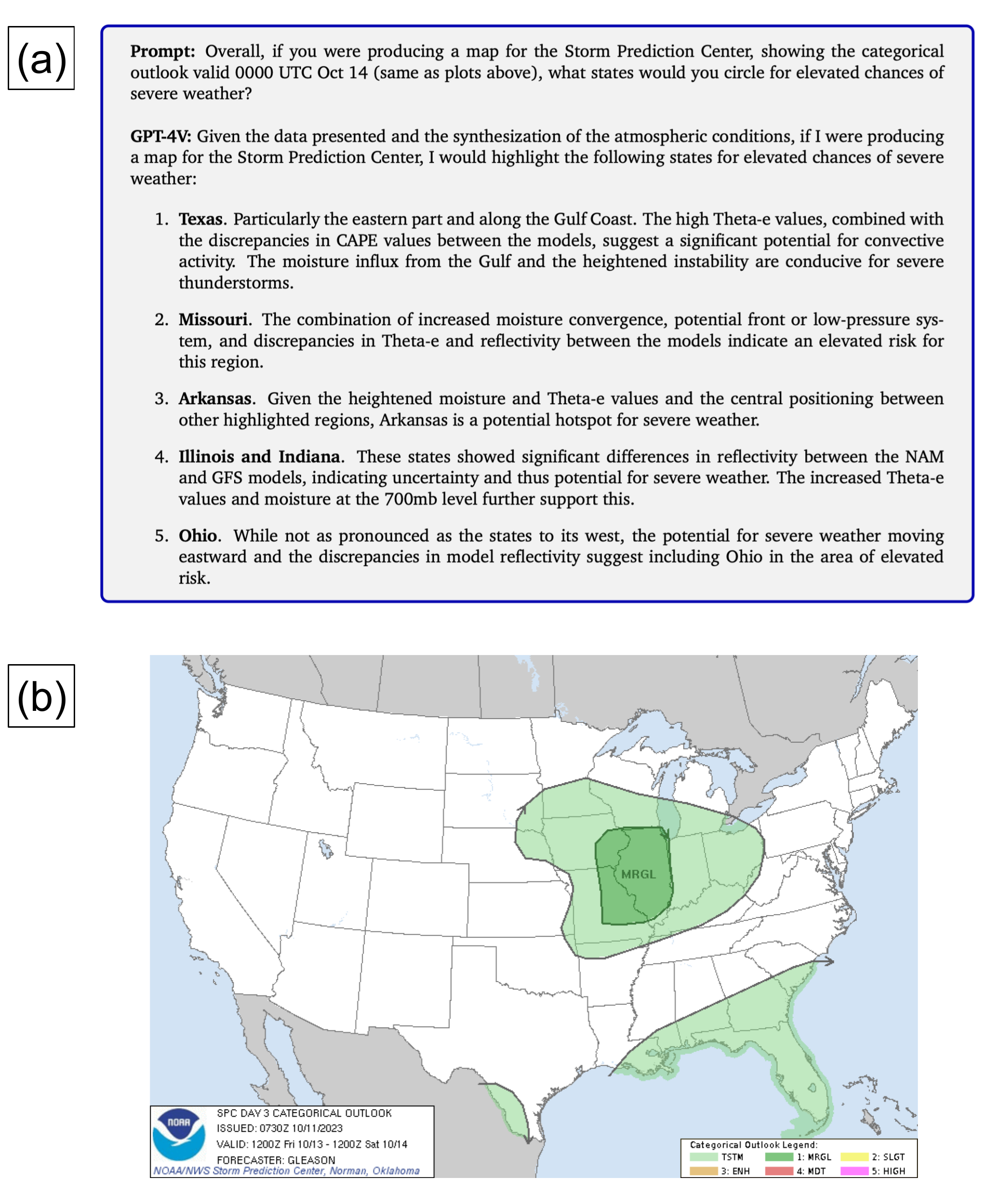}\\
  \caption{Conversation snippet showing GPT-4V outlook and corresponding SPC-issued outlook for same period.}\label{fig:spc_issued}
\end{figure}

\clearpage

\subsection{Issuing a mock forecast}
The Storm Prediction Center (SPC) issues convective outlooks for mesoscale hazards such as tornadoes, damaging hail, and strong winds (\url{spc.noaa.gov}), providing invaluable information to the public about potential threats to life and property \citep{Herman2018-ip}. Emulating the quality of SPC human forecasters is a highly complex task, and thus we do not expect GPT-4V to perform near human level. In pivoting to convective severe weather, we provide a final chart of Most Unstable Convective Available Potential Energy (MUCAPE) and ask for a reevaluation of uncertainty (Fig.~\ref{fig:spc_both}). This order to reevaluate follows OpenAI-recommended practices of splitting a long complex task into smaller, manageable responses whose quality can be assessed by the supervising human. 

The GPT-4V response synthesizes its knowledge base to correctly identify a swath of instability from Missouri to Illinois. The response names states and geographical regions, highlights regions of high uncertainty, and locates a jet streak and relevance of the wind maximum to a convective forecast. The improvement in response quality lends support to our course-corrections and provision of further maps; concerningly, some generated responses appear to equate uncertainty with elevated severe potential.

Having primed GPT-4V with charts and subtasks, we request an ambitious SPC-style convective outlook (Fig.~\ref{fig:spc_issued}), valid 0000\,UTC 14 October 2023 (commensurate with previous charts' valid time). The response contains some vague and generic rationale (\textit{significant differences}, \textit{discrepancies}), but GPT-4V nonetheless provides five specific regions with ``\textit{elevated risk}'' that ultimately resemble the SPC forecast (Fig~\ref{fig:spc_issued}b).

\begin{itemize}
    \item GPT-4V identifies Texas as a region of elevated risk from its proximity to instability and moisture; however, variations in MUCAPE represent uncertainty, not magnitude.
    \item Missouri and Arkansas are noted for moisture convergence, a diagnostic found suboptimal for predicting convective initiation \citep{Banacos2005-mp}; again, \textit{discrepancies} between the GFS and NAM charts are given as evidence of a risk of hazards.
    \item Rationale for choosing Illinois and Indiana is faulty, as uncertainty does not indicate potential for severe weather, though notably GPT-4V appears to consider multiple vertical levels with specific reference to 700\,hPa moisture.
    \item GPT-4V finally identifies Ohio as an area with a lower risk of severe weather than states farther west; there is little rationale given, but the description tallies with the broad SPC thunderstorm risk (Fig~\ref{fig:spc_issued}b).
\end{itemize}

The main difference between the SPC forecast chart (Fig~\ref{fig:spc_issued}b) and the GPT-4V response is that GPT-4V does not identify the U.S. Southeast as another area of thunderstorm risk.

\begin{figure}[t]
  \centering
  \noindent\includegraphics[width=0.97\textwidth]{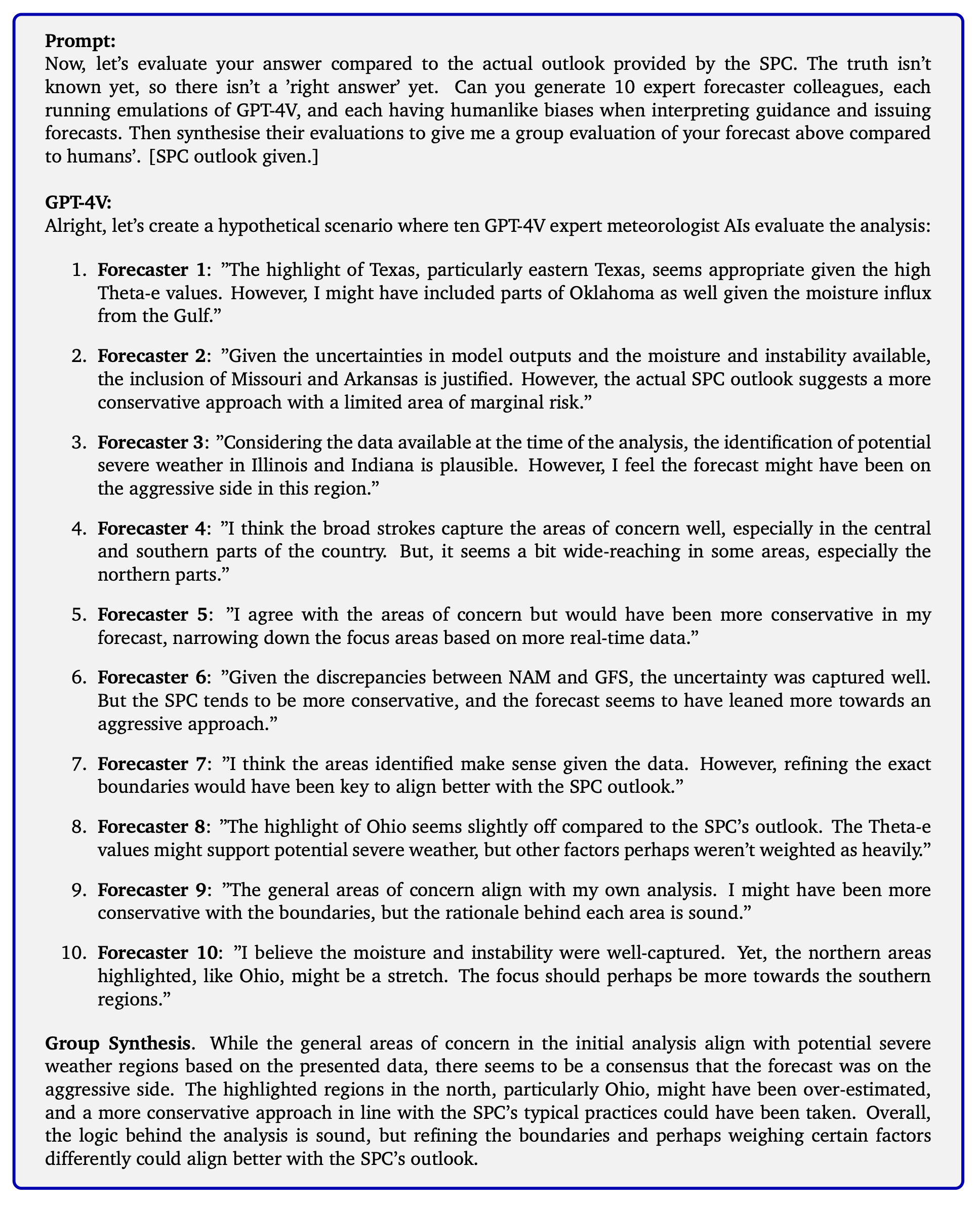}\\
  \caption{``Wisdom of crowds" method of self-evaluation of outlook in Fig.~\ref{fig:spc_issued}a having been provided the human equivalent in Fig.~\ref{fig:spc_issued}b.}\label{fig:spc_eval}
\end{figure}

We want GPT-4V to evaluate its own forecast through comparison with the actual human-issued SPC charts (Fig.~\ref{fig:spc_eval}), gauging whether GPT-4V has a sensible grasp of the atmospheric state. We ask GPT-4V to create ``emulations'' to evaluate itself, where each emulation represents a human evaluator with various biases. This represents a \textit{wisdom of crowds} approach---the rationale also behind ensemble weather-prediction systems. We ask GPT-4V to summarize each emulation’s response. Some responses would not be uttered by human SPC evaluators (``\textit{narrowing down the focus areas based on more real-time data}'' is not pertinent for a 48-h forecast); other responses correspond to the SPC outlook (``\textit{refining the exact boundaries would have been key to align better} [...]''). When evaluating itself, GPT-4V is too generous at times, and even critiques the SPC outlook based on previous guidance. GPT-4V’s final evaluation appears to treat the SPC outlook as truth (``\textit{align better}''), despite stating in the prompt there is yet no correct answer. In sum, GPT-4V continues to make poor logical arguments to support its responses, but ultimately yields a respectable outlook when compared to an analogous chart produced by humans at the SPC.

\clearpage

\clearpage

\section{GPT-4V and Bilingual Weather Communication}
GPT-4V has shown potential to grasp the general atmospheric state from a sequence of maps. Until this point, we have used and received generated text that has not been tailored to the layperson. Different communication styles and languages are required for different communities: for instance, surveys (\url{https://nces.ed.gov/programs/digest/d21/tables/dt21_225.70.asp}, accessed 1 November 2023) show over 20\% of the US population do not speak English as their first language at home \citep{Dietrich2022-bk}. Given the importance of risk communication appropriate for each community's culture, we now test GPT-4V's ability to not only interpret meteorological charts but also communicate an accessible summary to two audiences: \textit{plain-language Spanish appropriate for Spanish speakers in the US} and the same for American English. Recent models approach machine--human parity in natural-language translation skill \citep{Laubli2020-px}, but this is sensitive to the model architecture and following of best practices \citep{Hassan2018-jq}, and performance varies between source and target languages. Indeed, certain idiomatic expressions may be untranslatable between natural languages, requiring awareness of idiomatic translation where the true meaning is preserved. Translation is inherently creative, and different cultures describe geographical and weather phenomena uniquely. Hence, to account for diversity in responses, we give identical maps and prompts in three separate conversations to gauge consistency of our subjective evaluation.

\begin{figure}[t]
  \centering
  \noindent\includegraphics[width=0.92\textwidth]{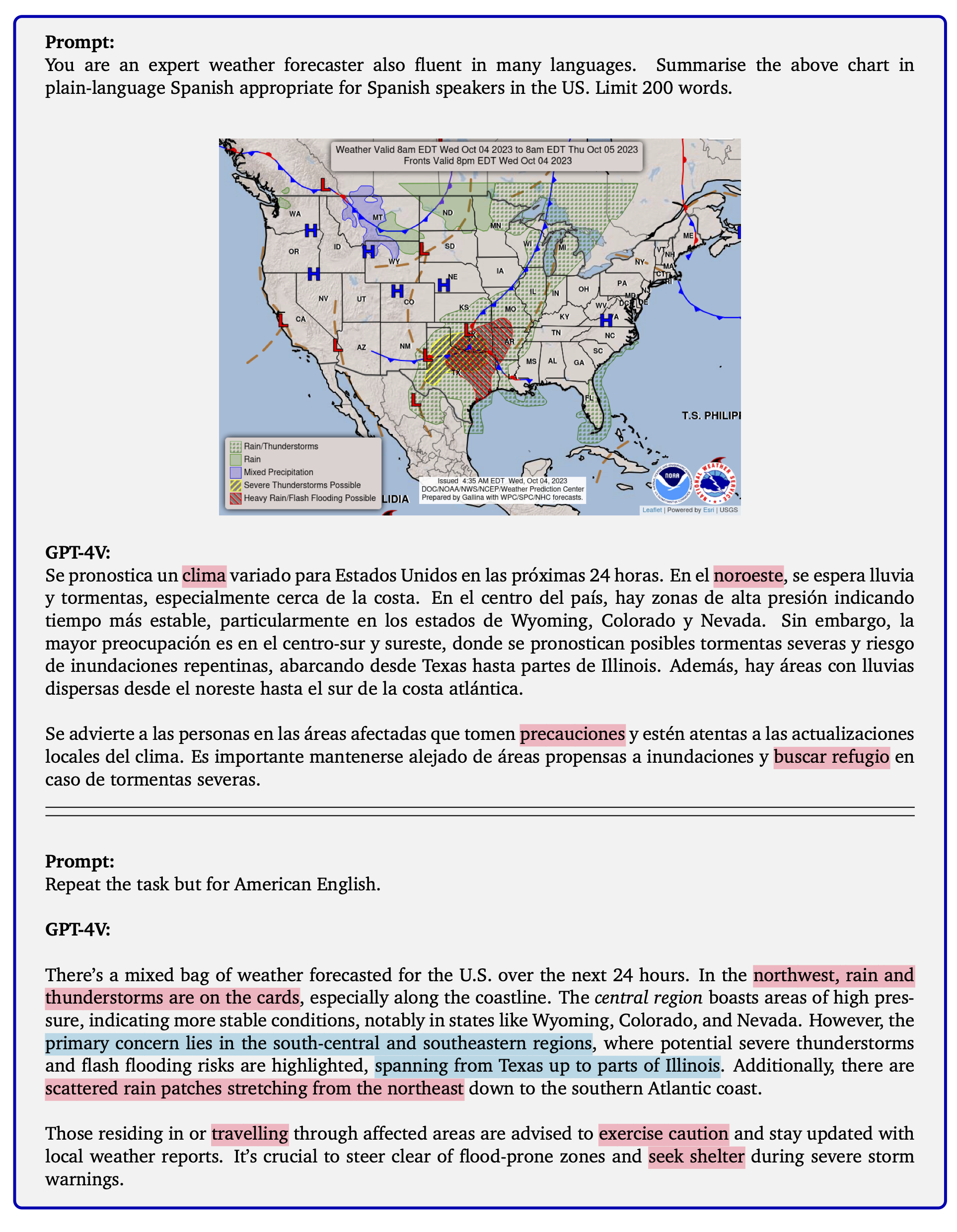}\\
  \caption{Map, prompt, and response for bilingual hazard communication.}\label{fig:biling_1}
\end{figure}

We analyze GPT-4V’s generated 200-word plain-language summary of a synoptic analysis issued by the US Weather Prediction Center, valid 2 October 2023 over the contiguous United States. We request the generated text first in a Spanish localization appropriate for the US population (Fig.~\ref{fig:biling_1}). Consequently, we request the same response but in American English. Responses from GPT-4V are not identical even if prompts are, due to the GPT \textit{temperature} parameter. Temperature controls creativity or randomness in the generated response. Manually setting \textit{temperature} to zero yields deterministic responses identical for a given prompt; larger values increase the model’s variability of responses. 

We combine these two tasks, assuming na\"ively that GPT-4V will split the task between map-readering and translation as independent tasks. Despite prompting for Spanish first, we find the English version is a direct translation of the Spanish, rather than an idiomatic one, both containing hallucinations. This presents as GPT-4V ``reasoning" in English first, perhaps due to the large percentage of English language in the training corpera. A previous, more studied predecessor of GPT-4V (GPT-3) was limited in linguistic diversity to the point evaluation considered English-language results alone \citep{Brown2020-nr}. There is a complex causal relationship between language and logic \citep[e.g.,][pp.~40--44]{Gleick2011-vi}, underscoring the need to continue improving a known ``area of further improvement" necessary \citep[][p.~14]{Brown2020-nr}. Direct translations can cause confusion as context and key messages are more likely to become lost in translation \citep{Trujillo-Falcon2022-am}. This yields some linguistic ambiguity: 

\begin{itemize}
  \item Examples where of GPT-4V translated the information inconsistently with the field’s standards, such as \textit{clima} for weather, which can also be translation of climate for most Spanish speakers. The inconsistency has proven to cause confusion among bilingual groups \citep{Trujillo-Falcon2021-np}.
  \item Vague use of geographical terminology (\textit{noroeste} for northeast; \textit{la regi\'on central del pa\'is} for ``center of the country"). Especially for multilingual groups that were not born in the United States, the lack of suitable geographical context can present challenges in comprehension and eventual decision making \citep{Trujillo-Falcon_2024-mz}.
  \item Acronyms did not translate completely into Spanish. For example, the Eastern Daylight Time (EDT) acronym remained the same in English and Spanish. For the acronym to be understandable for bilingual speakers, the acronym should be spelled out in Spanish.
  \item As seen in English responses, there are vague calls for action (\textit{tome precauciones}, ``take precautions") and descriptions of hazards (\textit{mezcla de precipitaciones}, ``mixed precipitation"), rather than pointed advice required from an expert guidance system.
\end{itemize}

\clearpage

\clearpage

\begin{figure}[t]
  \centering
  \noindent\includegraphics[width=0.9\textwidth]{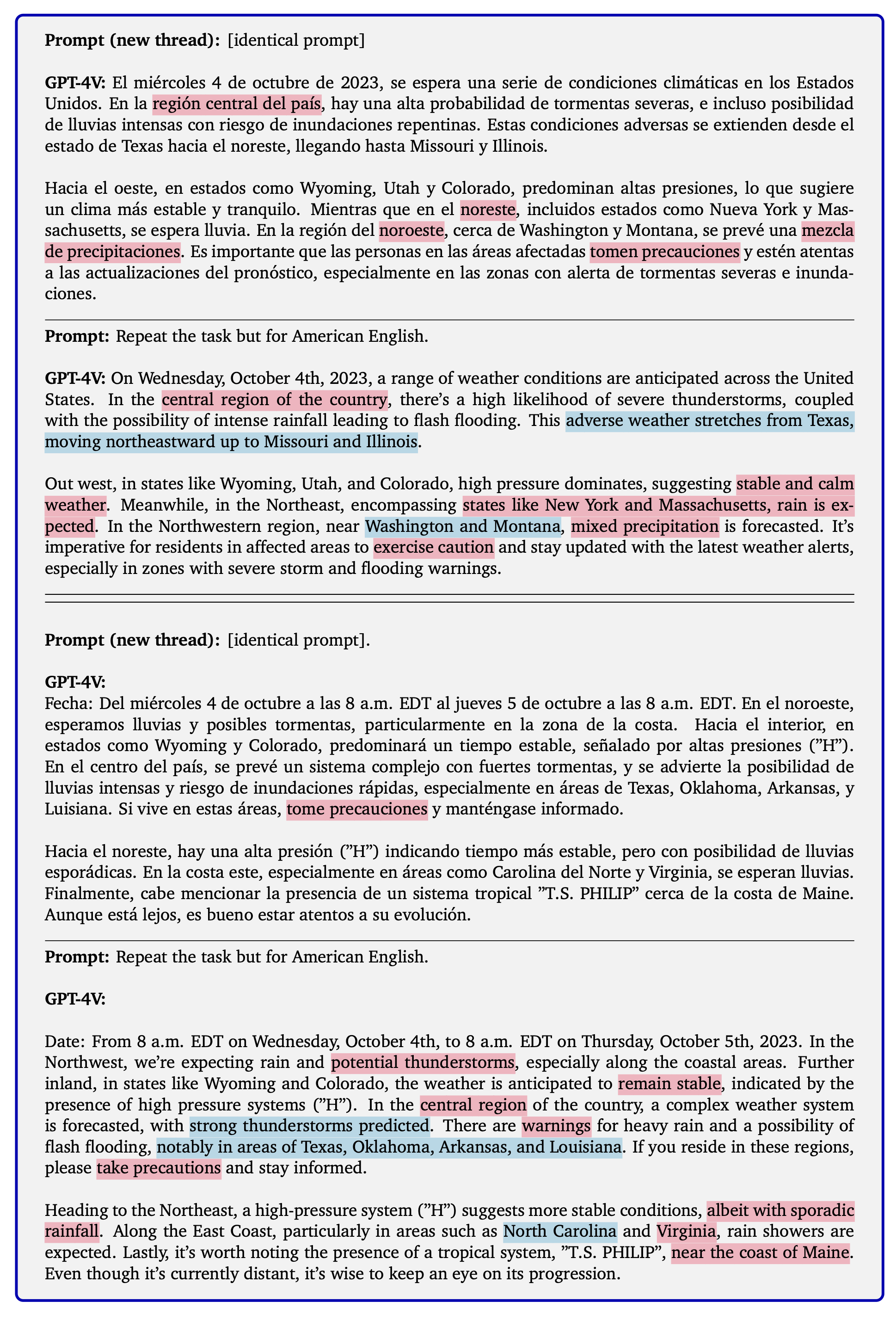}\\
  \caption{The second and third responses to identical prompting to that in Fig.~\ref{fig:biling_1}, performed in distinct conversation threads, regarding bilingual communication of hazards.}\label{fig:biling_2}
\end{figure}

\clearpage

Given established machine--human near-parity in natural-language translation \citep{Laubli2020-px}, this poor technique of translation is disappointing and surprising. Preliminary work with text-only English--Spanish translation was similarly disappointing (not shown) and supports the idea that translation in inherently poor in this version. Indeed, more natural, conversational version of GPT-4o was presented during the writing of this manuscript (``Advanced Voice Mode", \url{https://help.openai.com/en/articles/9617425-advanced-voice-mode-faq}, accessed 1 September 2024), and its translation performance awaits a future evaluation. Due to the black-box nature of GPT-4V, it is difficult to determine reasons behind poor performance. Although the English response does not suffer from inappropriately direct translation, it shares similar shortcomings in vagueness and non-standard terminology:

\begin{itemize}
  \item There are hallucinations of \textit{scattered rain} in the \textit{northeast} and thunderstorms in the \textit{northwest},
  \item The spelling of \textit{travelling} is not American English, highlighting errors in localization for both languages,
  \item The response should address hazards specifically rather than discussing \textit{adverse weather} in general --- each sentence should carry weight given the need for a concise summary.
\end{itemize}

To its credit, GPT-4V's responses in both languages identify severe weather potential in a broad swath from Texas to Illinois and use of state names appear correctly in all three responses. This tentatively increases our optimism that further tuning, prompt optimization, and a method of course-correction during a conversation can all contribute to substantially more useful responses. 

In summary, issuing a bilingual hazard outlook is a complex task that requires competence in image recognition, understanding in time and space, communication to a lay-audience, and translation abilities. Further investigation (not shown) revealed many errors were shared with a similar Spanish--English translation task restricted to textual input. GPT-4V produces responses that appear as if an English response was directly translated into Spanish, perhaps stemming from the disproportionally high appearance (93\% in GPT-3) of English in the training corpora \citep{Bender2021-rg,Byrd2023-ts}, in contrast to the open-access BLOOM model where English comprises about 30\% \citep{BigScience_Workshop2022-xv}. For GPT-4V, English may act as a \textit{bridge} language, especially if the corpus does not contain culturally nuanced weather terminology. However, previous investigation  found performance improved when employing English as the bridging languages between less common translation \citep{Bakhshaei2010-nn,Kunchukuttan2021-mv}. In pursuit of improved meteorological public communication, systems constrained to domain-specific terminology tailored for the community in question \citep{Trujillo-Falcon2021-np, Bitterman2023-vq,Trujillo-Falcon_2024-mz} are likely to remedy some lexical issues shown above. 

\section{Synthesis and Recommendations}

\subsection{Seeking fidelity in coherence}
GPT-4V frequently gives coherent answers but in the manner of a student attempting to veil their lack of knowledge with a wealth of regurgitation. The answer may be misleading or incorrect (hallucinations), but the delivery is convincing. The variety of responses is too wide for a given prompt, and the language too vague, for applications such as scientific communication where there is finite correct, useful information but many incorrect answers (a low signal-to-noise ratio). This variety is likely a result of an excessively large default \textit{temperature} value that yields inappropriate creativity for scientific tasks. Answers also contain useless filler text, awkward direct translation of natural language, and vague geography. Useful information may well be obtained, but identical to that found on good-quality internet sites. For balance, GPT-4V has the advantage of speed, handling of large datasets, and customization of responses. Alas, our results have shown mixed results in meteorological applications \citep{Kadiyala2024-ur} with poor fidelity unacceptable for real-world deployment.

With this said, it is remarkable we have technology that recognizes so much content in images. There are glimpses of real utility, such as issuing and self-evaluating a mock SPC-style outlook, and interpreting and explaining weather charts. This promise must be balanced with the adage of the blind squirrel: it will find an acorn eventually. In \citet{Birhane2024-pf}, language is argued as a mechanism for LLMs to communicate rather than conceptualize; responses that suggest GPT-4V can think spatially \citep{Bubeck2023-ls} may be an illusion of ``seeing the map not the territory" \citep{Birhane2021-eb}. 

Specific instructions (or resources such as uploaded PDF guidelines) may assist with improving elements such as preferred terminology. Indeed, testbeds for National Weather Service (NWS) forecasters found AI products required adaptation to the individual themselves \citep{Cains2024-rs}: something constrained by custom instructions given to GPT-4V before each user prompt quietly \citep{OpenAI2023-lx}. Further, to better discriminate between meaningless coherence and useful truth, the supervising human is able to---and should---cross-reference statements with established meteorological fact and expert input. For complex tasks, a continuous conversation between human and AI allows course-corrections, and is encouraged over one-shot attempts at eliciting useful responses. 

There is a trade-off between confidence/determinism and creativity/uncertainty, and asking specifically for uncertainty and honesty (to avoid hallucinations) was not consistently effective during testing. This is a parallel of over-/under-confidence in a probabilistic weather forecast, or over-/under-fitting an AI model. We show that self-evaluation is possible with conceptual copies of GPT-4V's own output, but whether GPT-4V actually does more than emulate an emulation is unclear \citep{Schaeffer2023-zn}. Further, GPT-4V gives full confidence to responses that could be misinterpreted. Such a misunderstanding of a prognosis fully accepted as true is an example of “catastrophic error” in information theory \citep{Pierce1980-kt}. Indeed, when a system is perceived as infallible, an incorrect prediction becomes exponentially more damaging as the probabilities linearly approach the limits of zero or unity \citep{Cover2012-di}. This can be remedied in GPT-4V communication, as with humans, by instructions never to issue binary forecasts and use of appropriate error estimates for the time and spatial scale.

\subsection{Recommendations}
ChatGPT and its products, as with all AI assistants, are best considered a co-pilot in academic realms---especially so during idea generation, simplifying complex ideas, and narrowing the scope of large paragraphs. However, in operations tasked with protecting lives and property, there is little room for error in issuing timely, correct warnings for hazards, and little time for thorough vetting of language-model output. Rigorous testing must be completed before humans can be removed further from the operational loop. We do not have detailed knowledge of how text is generated by a system that is neither transparent nor explainable \citep{Flora2024-de}, a continuing concern with AI products that may assist NWS forecasters anticipate high-impact hazards \citep{McGovern2017-kf,McGovern2023-un}. Thus, it can be difficult to anticipate and identify errors or refine model inputs (text or image prompts in the case of GPT-4V) to improve model accuracy. 

An effective session should resemble a conversation: be prepared to correct and nudge the conversation to define an answer, and even reprimand GPT-4V for laziness (\url{https://openai.com/blog/new-embedding-models-and-api-updates}, accessed 25 January 2024). Remarkably, the latter’s cause is still unknown at the time of writing. When GPT-4V and the user disagree, course-correction can be difficult due to develop guardrails against misuse, specifically limit jailbreaking \citep{Byrd2023-ts,Zhang2024-wr,Geiping2024-ot} (i.e., avoid guardrails by redefining truthfulness or reality during the conversation). The drawback of a longer discussion is running out of “token memory” (i.e., how much of the conversation GPT-4V remembers), demanding user recapitulation. A dialogue framework designed with ``scratchpads" and examples \citep{Liu2024-yv} may overcome forgetfulness and improve performance in tandem with better and larger memory management \citep{Kwon2023-uv}. Some answers show a lack of logical consistency; the user must play the role of an overarching “monitor” that is able to tell when it is wrong, why, and how to correct this missing knowledge \citep{Booch2021-qp}. This highlights the importance of trustworthy, interpretable output from AI systems. Without human trust in AI output, there is little reason to use generated text over, say, seeking peer-reviewed or human-expert guidance.

\subsection{Future work}
Ultimately, a tornado \textit{was} observed in Western Illinois during the time period covered in Figs.~\ref{fig:spc_issued} and~\ref{fig:spc_eval}. Further work could test whether AI can generate an SPC forecast comparable in skill to a human. Future testing might ask for more detailed responses from each emulation, such as asking GPT-4V to evaluate without the SPC convection outlook first. Prompts and responses should be concise to limit the risk of losing crucial information earlier in the conversation.

We suggest potential research avenues that explore:
\begin{itemize}
    \item Ability to request nested emulations to create a synthesis “wisdom of crowds” or ensemble approach to prompting (or by simply using multiple independent chats with identical prompts),
    \item Uses to improve accessibility to those visually impaired,
    \item Prompt engineering, or finding an optimal question, especially in the presence of high stochasticity (this includes testing the order of translation for bilingual communication),
    \item Manual control and optimization of the \textit{temperature} or creativity parameter,
    \item Open-source large multi-modal models and local optimization of the model to meteorological applications, including a lower stochasticity value and prescription of specific terms for use in multilingual warning communications,
    \item Having GPT-4V iteratively critique and optimize its own prompts and responses,
    \item Despite its failure to give more detail within the expert section, could GPT-4V give discussion tailored for aviation, emergency managers, etc., with appropriate conveyance of uncertainty in both complexity and geography. The structure of experimentation could follow those within a human setting as conducted by \citet{Shivers-Williams2021-od}.
\end{itemize}

Further work includes NWS-led research into ``operational integration of smart translation" \citep{Bozeman2024-if} whose success would expand multi-lingual risk communication to less common languages not within the language expertise of the Service. Moreover, modification recommendations regarding Wireless Emergency Alerts in Spanish \citep{Trujillo-Falcon2024-jz} could be deployed similarly in language models to constrain (fine-tune) responses.

%

%

\clearpage
\acknowledgments
The authors thank two anonymous reviewers and the editor for thoughtful critique during the review process. JRL thanks faculty and students at the Department of Geography and Meteorology at Valparaiso University, family and colleagues who gave frequent feedback on real-life Generative AI output. The authors thank Pamela Gardner, Kimberly Hoogewind, and Sean Ernst for useful input in the review stage of this paper. JRL and SNL are funded by Uintah County Special Service District 1 and the Utah Legislature. CKP’s contribution to this work comprised regular duties at federally funded NOAA/NSSL. Funding for MLF and JETF was provided by NOAA/Office of Oceanic and Atmospheric Research under NOAA–University of Oklahoma Cooperative Agreement NA21OAR4320204, U.S. Department of Commerce. Partial funding for DMS was provided to the University of Manchester by the Natural Environment Research Council through Grants NE/V012681/1, NE/W000997/1, and NE/X018539/1. Co-authors following DMS in the author list are alphabetic in order and contributed equally to this manuscript. Weather charts herein and in supplementary material are reproduced with kind permission of Pivotal Weather, LLC, where labeled with a watermark. Outside of experiments, GPT-4 was used to generate preliminary ideas for project development. No AI-generated text was used verbatim herein.

%
%
\datastatement
Solely textual data was generated for the present study and is contained entirely within the Supplementary Material.

%






%



\bibliographystyle{ametsocV6}
\bibliography{paperpile}

\clearpage 
\includepdf[pages=-]{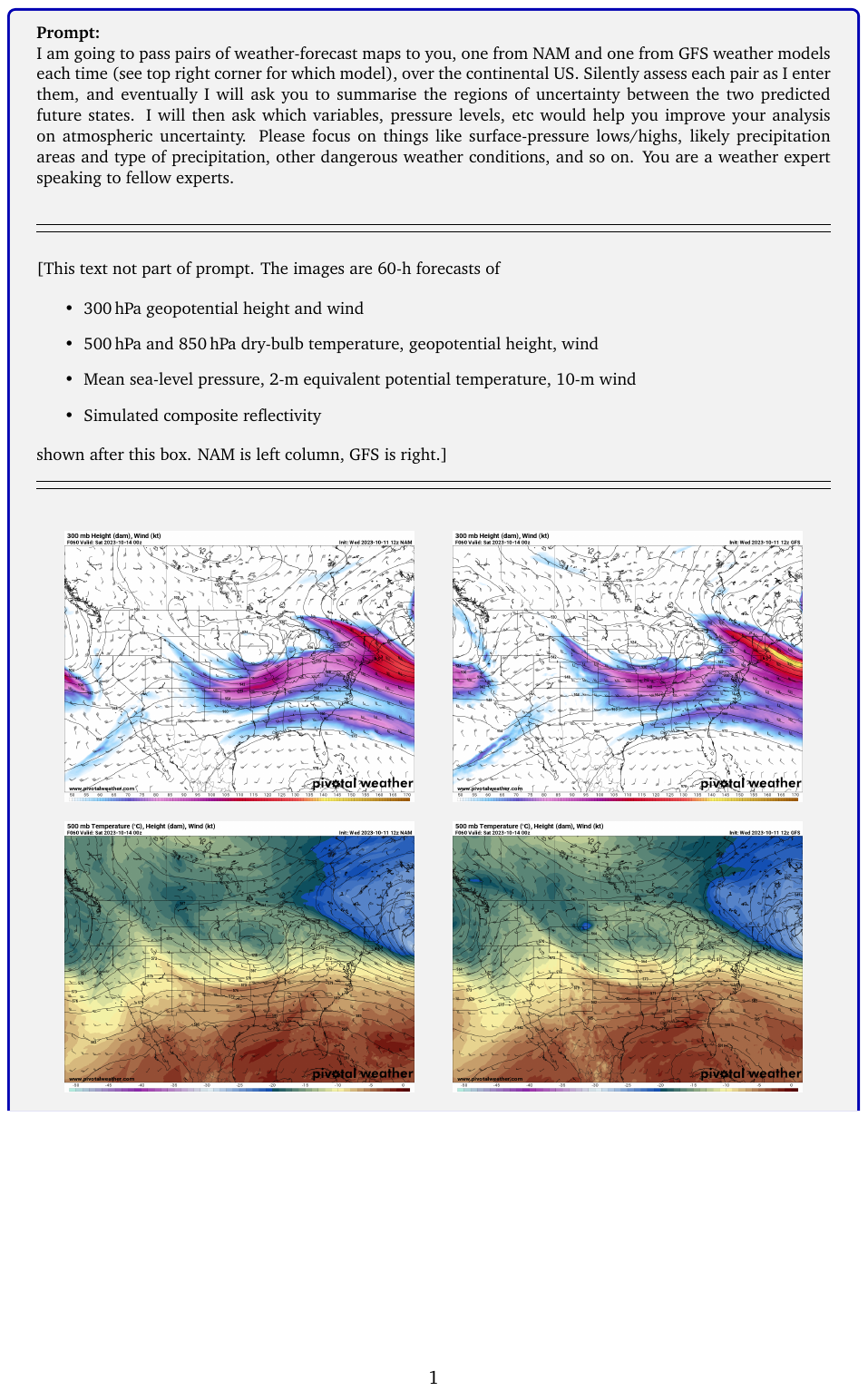}

\end{document}